\documentclass[a4paper,UKenglish]{dagrepdigitalarchaeoludology}

\usepackage{microtype}
\usepackage[utf8]{inputenc}

\usepackage{xcolor}
\usepackage{soul}
\usepackage{caption}
\usepackage{subcaption}
\usepackage{upgreek}

\usepackage{apacite}

\usepackage{booktabs}
\usepackage[final]{pdfpages}

\usepackage{csquotes}

\newcommand{\reffigure}[1]{Figure~\ref{#1}}
\newcommand{\refsection}[1]{Section~\ref{#1}}
\newcommand{\refsubsection}[1]{Subsection~\ref{#1}}
\newcommand{\reftable}[1]{Table~\ref{#1}} 

\newcommand{\DA}{Digital Arch\ae oludology}
\newcommand{\LUDII}{Ludii}

\subject{Report from Dagstuhl Research Meeting 19153}
\title{Foundations of \DA} 

\hypersetup{
pdftitle={Foundations of Digital Archaeoludology},
pdfsubject={Digital Archaeoludology},
pdfauthor={Cameron Browne, Dennis J. N. J. Soemers, \'Eric Piette, Matthew Stephenson, Michael Conrad, Walter Crist, Thierry Depaulis, Eddie Duggan, Fred Horn, Steven Kelk, Simon M. Lucas, Jo\~ao Pedro Neto, David Parlett, Abdallah Saffidine, Ulrich Sch\"adler, Jorge Nuno Silva, Alex de Voogt, and Mark H. M. Winands},
pdfkeywords={Digital Archaeoludology, games, board games, abstract games, history, archaeology, artificial intelligence}
}

\titlerunning{19153 -- \DA}

\author[1]{Cameron Browne}
\author[1]{Dennis J. N. J. Soemers}
\author[1]{\'Eric Piette}
\author[1]{Matthew Stephenson}
\author[2]{Michael Conrad}
\author[3]{Walter Crist}
\author[4]{Thierry Depaulis}
\author[5]{Eddie Duggan}
\author[4]{Fred Horn}
\author[1]{Steven Kelk}
\author[6]{Simon M. Lucas}
\author[7]{Jo\~ao Pedro Neto}
\author[4]{David Parlett}
\author[8]{Abdallah Saffidine}
\author[9]{Ulrich Sch\"adler}
\author[7]{Jorge Nuno Silva}
\author[10]{Alex de Voogt}
\author[1]{Mark H. M. Winands}
\affil[1]{Department of Data Science and Knowledge Engineering, Maastricht University, the Netherlands
  \texttt{cameron.browne@maastrichtuniversity.nl}}
\affil[2]{Kunsthistorisches Institut, Universit\"at Z\"urich, Switzerland}
\affil[3]{Department of Anthropology, American Museum of Natural History, United States of America}
\affil[4]{Independent}
\affil[5]{Department of Science, Technology and Engineering, University of Suffolk, England}
\affil[6]{School of Electronic Engineering and Computer Science, Queen Mary University of London, England}
\affil[7]{University of Lisbon, Portugal}
\affil[8]{School of Computer Science and Engineering, University of New South Wales, Australia}
\affil[9]{Swiss Museum of Games, Switzerland}
\affil[10]{Economics \& Business Department, Drew University, United States of America}
\authorrunning{Cameron Browne {\it et al.}}

\keywords{Games, \DA, Artificial Intelligence, Data, Archaeology, Anthropology, History, General Game Playing, Ludemes, Game Modelling}

\seminarnumber{19153}
\semdata{10.--12.~April, 2019 -- \url{https://www.dagstuhl.de/19153}}
\subjclass{
            H.1.2 User/machine Systems,
            H.2.8 Database Applications,
            H.3.3 Information Search and Retrieval,
            H.3.5 Online Information Services,
            H.3.7 Digital Libraries,
            I.2.1 Applications and Expert Systems,
            I.2.4 Knowledge Representation Formalisms and Methods,
            I.2.6 Learning,
            I.2.8 Problem Solving, Control Methods, and Search,
            I.6.8 Types of Simulation,
            J.5 Arts and Humanities,
            K.3.1 Computer Uses in Education}

\volumeinfo
  {Cameron Browne, Dennis J. N. J. Soemers, \'Eric Piette, Matthew Stephenson, Michael Conrad, Walter Crist, Thierry Depaulis, Eddie Duggan, Fred Horn, Steven Kelk, Simon M. Lucas, Jo\~ao Pedro Neto, David Parlett, Abdallah Saffidine, Ulrich Sch\"adler, Jorge Nuno Silva, Alex de Voogt, Mark H. M. Winands}
  {18}
  {Foundations of \DA}
  {}
  {}
  {1}

\begin{document}

\maketitle


\begin{abstract}

\DA\ (DAL) is a new field of study involving the analysis and reconstruction of ancient games from incomplete descriptions and archaeological evidence using modern computational techniques. 
The aim is to provide digital tools and methods to help game historians and other researchers better understand traditional games, their development throughout recorded human history, and their relationship to the development of human culture and mathematical knowledge. 
This work is being explored in the ERC-funded Digital Ludeme Project.

The aim of this inaugural international research meeting on DAL is to gather together leading experts in  relevant disciplines -- computer science, artificial intelligence, machine learning, computational phylogenetics, mathematics, history, archaeology, anthropology, etc. -- to discuss the key themes and establish the foundations for this new field of research, so that it may continue beyond the lifetime of its initiating project. 

\end{abstract}


\tableofcontents









\section{Introduction} \label{Sec:Introduction}

\presentations{Cameron Browne}


Based on research in areas such as archaeology, anthropology, and history, it is well known that humans across many cultures played a wide variety of games throughout history. An improved understanding of what games were played, and how they were played, in different cultures, locations, or moments in time, can also lead to other insights in these research fields (e.g. sometimes shared games can be evidence of likely contact between cultures). Evidence found typically only paints a partial picture of how games were played historically \cite{Murray1951, Shadler1998Mancala, Shadler2013GamesGreekRoman}. For example, evidence can consist of boards and pieces without recorded rules, or incomplete written descriptions of a rule set. Establishing a more complete view may require combining many different pieces of evidence, but often is impossible to do with 100\% certainty.

In this report, and its associated Dagstuhl research meeting, we introduce the research field of \DA\ (DAL) as \textit{``the use of modern computational techniques to improve our understanding of the development of traditional strategy games''}. The ERC-funded Digital Ludeme Project\footnote{\url{http://www.ludeme.eu/}} (DLP) \cite{Browne2018ModernTechniques} marks the start of this research field, but we hope that the field will continue to be active beyond the duration of the project. 

To facilitate the use of modern computational and data-driven techniques, we use a general ludeme-based approach that can model all of the included games in a single, easily-comprehensible game description language. Games are modelled as trees of ``ludemes'' \cite{parlett16}, which can intuitively be understood conceptual units of game-related information.\footnote{Appendices A and B contain previously unpublished notes by Thierry Depaulis that shed some light on the origins of the term ``ludeme''.}

While computational techniques cannot provide a substitute for manual discovery and analysis of evidence through field work, they can provide automated analysis of existing evidence at a significantly larger scale and speed than manual work. This may provide useful insights or generate plausible new hypotheses that can be further investigated by field experts. 
For example, if a choice is to be made between possible interpretations of given piece of evidence, then mathematical analyses performed by our system may help the investigator decide which of the possible options is the more likely.
In addition, we expect the research to lead to valuable contributions to the Artificial Intelligence (AI) research community, such as the new ludeme-based game description language and related tools.


\subsection {Scope of Games}

Defining the scope of games to be covered in this study requires considerable care. 
We describe the games of interest as ``traditional strategy games'' and understand these to be games that:

\begin{itemize}
    \item have no proprietary owner, i.e. exist in the public domain,
    \item show some longevity or historical relevance to the culture with which they associated,
    \item reward mental skill rather than physical skill (i.e. no sports), and
    \item in which players must make decisions, with good decisions generally outperforming bad decisions.
\end{itemize}

These are not necessarily hard requirements for a game to be included in the project, but are general guidelines. 
This class of games includes mostly board games, some card games, tile games, dice games, etc., and the non-proprietary requirement means that most games of interest will be pre-industrial, i.e. invented before circa 1875, although there may be exceptions.

The main criteria for inclusion in this study is a game's impact on the evolutionary record; as we aim to compute as complete a picture of the development of traditional strategy games as possible, we wish to include the key {\it influencers} that might have shaped this process. 
We will therefore be including important outliers in the study that do not exactly fit our criteria listed above -- such as the important family of Snakes \& Ladders games which do not involve any element of strategy -- if their inclusion helps shed light on the development of traditional strategy games.

The game Surakarta is an interesting edge case that will test our criteria for inclusion in the study. 
Bell carefully states that: ``{\it Surakarta takes its name from the ancient town of Surakarta in Java}''~\cite[p.32]{Bell1973}. 
However, while this game is often described as {\it the} Javanese board game, there is no evidence that the game originated in Java or is even played there; there is suggestion that it may be a recently {\it invented tradition}\footnote{https://boardgamegeek.com/thread/389921/fun-not-just-topologists} whose true origins lie in Korea or China.\footnote{https://web.archive.org/web/20131029192809/http:/www.encykorea.com/Contents/play/child04.htm}

We acknowledge the dangers of using the term ``traditional'' in describing those games on which we are focussed. 
This is a commonplace word that has taken on technical meaning in the field of anthropology in particular, where its exact meaning has been debated for decades~\cite{Shanklin1981} and there is no sign of a universally accepted definition emerging any time soon. 
We do not intend to enter this debate, but instead use this term in its broader non-technical form as describing artefacts or practices that are long established or have some relevance to tradition. 

It is also worth clarifying our usage of the term ``strategy game'' in this context, to avoid confusion. 
We include games that reward mental skill over physical skill and in which good decisions generally beat bad decisions. 
We are not using the hard definition of {\it pure strategy} games from Combinatorial Game Theory (CGT), but instead refer to any game that involves at least some element of strategic decision making. 
This includes games with chance elements, hidden information, simultaneous moves, etc., provided that players make at least some decisions that affect the outcome of the game. 
This resonates with the implementation of games in our associated Ludii general game system as sequences of {\it decision moves} by players followed by forced consequences. 

In this computational study, we must mathematise games in order to model them in a single consistent format. 
We therefore treat games as mathematical entities, and as cultural artefacts consisting of:

\begin{enumerate}
    \item {\it evidence} in the form of physical remains, and
    \item {\it ideas} in the form of rules.
\end{enumerate}

The social dimension of games as ``play practice'' is harder to quantify, e.g. the role of games as social lubricants~\cite{Crist2016} or as spectacle for onlookers, especially as historical evidence in this respect is scarce and must often be deduced from modern observations. 
We will incorporate this aspect into our analysis where possible and appropriate. 

It is worth emphasising that the aim of DAL is {\it not} to identify a single ``correct'' rule set for any given game. 
We expect the software tools that we are developing to yield a distribution of plausible rule sets for each game, with estimates for game quality and historical authenticity, for the user to interpret as appropriate. 
Clustering techniques may be applied to identify the archetypal or ``average'' rule set within a cluster of similar variants, which may still yield useful insights regarding similarities between games without any guarantees of historical accuracy. 

Further, while this study will focus on strategy games, we will also include other important games that might have had some impact on their development throughout history. 
For example, any study of the history of board games would not be complete without Snakes \& Ladders, even though it does not fit the profile of the games we are primarily interested in.
A core aim of this study is to develop as complete a picture as possible of the development and dispersal of traditional strategy games, so we will be including representative examples of this category (according to our definition) plus close relatives which might have had some impact upon them.

While we strive to include as many cultures, locations, and periods of time as possible in our analyses, it is important to acknowledge that it will only be possible to cover cases for which data is available. There will inevitably be biases in the availability of data, for instance due to certain regions being safer, or otherwise more convenient or preferred for field work, than other regions.

\section{Prior Research in Games and Playing Practices Throughout History}

In this section, we summarise a number of presentations and discussions about various aspects of prior research in games and playing practices throughout history. This focuses primarily on research fields such as archaeology, anthropology, history, etc. The subsequent section similarly summarises a variety of computational and data-driven techniques, and -- taking the conclusions of this section into account -- discusses how they may be useful for further research in these more traditional areas of research.


\abstracttitle{Game Classifications} \label{Subsec:GameClassifications}

\presentations{Thierry Depaulis and David Parlett}


There have been numerous attempts at constructing comprehensive classifications of games. One of the oldest known classifications of games dates back to 1283 \cite{AlfonsoX1283}, in which games are split up as follows:
\begin{enumerate}
    \item Games that are played on horseback.
    \item Games that are played dismounted (fencing, wrestling, etc.).
    \item Games that are played seated.
\end{enumerate}
Of these three classes, only the third is relevant for this report. This class was further split up into the following categories:
\begin{enumerate}
    \item Games that rely ``on the brain''.
    \item Games that rely only on chance.
    \item Games that ``take the best of both''.
\end{enumerate}
Of these categories, the first and the third are relevant for this report, but the second -- which would, for instance, include games of which the outcome is determined purely by dice rolls -- is not.

In the game AI community \cite{yannakakis2018artificial}, games are often split up in categories according to whether they are deterministic or nondeterministic, whether players have perfect or hidden information, whether they are single-player or two-player or multi-player games, whether they are simultaneous-move or turn-based games, etc. This could be viewed as an extension of the three categories described above. The properties that these categories are based on are often mentioned in game AI research because they generally determine which AI algorithms may or may not be applicable, but further distinctions into more fine-grained classes are rare in AI literature.

\citeA{Murray1951} and \citeA{Parlett1999OxfordHistory} describe classification systems that classify games according to the general goals that players aim to accomplish in games -- war games, racing games, etc. During this Dagstuhl research meeting, Parlett noted that a classification system that also takes into account the mechanisms of how a game is played would be preferable.

Thierry Depaulis proposes a classification scheme based on seven features; \textbf{level of determination}, which accounts for (non)determinisim and (im)perfect information, the \textbf{main objective} (being the first to reach a position, capturing certain pieces, etc.), the \textbf{balance of forces} (symmetric or asymmetric), the \textbf{nature of pieces} (identical pieces, differentiated pieces, etc.), the \textbf{moves} (regular moves, sowing, etc.), \textbf{conflict resolution} (no conflict, pieces sent back to start, pieces removed altogether, etc.), and the \textbf{method of capture} (replacement, leaping, approach or withdrawal, etc.). This leads to three main classes, with some sub-classes, listed in \reftable{Table:ThierryClassification}.

\begin{table}[h]
\caption{Main classes and sub-classes proposed by Thierry Depaulis.}
\label{Table:ThierryClassification}
\centering
\begin{tabular}{@{}ll@{}}
\toprule
     & Examples \\
     \cmidrule(lr){2-2}
     \textbf{1. Simple Race Games} (pure chance, 1 piece per player) & Game of the Goose \\
     \cmidrule(lr){1-2}
     \textbf{2. Complex Race Games} & \\
     Taken pieces re-entered from start & Caupad/Causar, Pachisi \\
     Taken pieces temporarily immobilised & Backgammon \\
     Taken pieces eliminated & T{\^a}b \\
     \cmidrule(lr){1-2}
     \textbf{3. Games of Pure Intellectual Skill} & \\
     Games of traversal & Bair, Chinese Checkers \\
     Games of alignment & Merels \\
     Blockade games & Mu Torere, Haretavl \\
     Sowing games & Mancala \\
     Games of symmetrical elimination & Draughts \\
     Games of asymmetrical elimination & Fox and Geese \\
     Games of selective capture, symmetrical & Chess \\
     Games of selective capture, asymmetrical & Hnefatafl \\
     Games of territorial contest & Go \\
\bottomrule
\end{tabular}
\end{table}

Many different classification systems for games are possible, based on different types of features, which likely lead to very different classifications. Games may be classified according to rules (or ludemes used), based on more broad properties such as determinism vs. nondeterminism, based on general goals, based on origin, etc. Different systems will likely have different advantages and disadvantages for different purposes or target audiences (players, historians, AI researchers, etc.).

Alternatively, it may be the case that our analysis of the corpus of traditional strategy games produces some new classification scheme, based on fundamental relationships between their component rule sets and their historical and cultural context. 
This could be a useful outcome of the study. 


\subsection{Corpus and Data Collation} \label{SubSec:CorpusAndDataCollation}

\presentations{Walter Crist and Alex de Voogt}

In \refsection{Sec:Introduction}, we attempted to establish guidelines for which games should or should not be included in the analyses of the project. These cover most of the cases we are interested in, but would exclude some games which we still consider to be important. For instance, the game of \textit{Snakes and Ladders} would be excluded due to a lack of strategic choices, but is still relevant to include due to its influence and cultural importance. Additional criteria that could be considered for game inclusion, which would cover the case of Snakes and Ladders, would be to include games that are (or were at some point in time):
\begin{itemize}
    \item influential, or
    \item unique, or
    \item representative (of a time, a location, a culture, etc.), or
    \item widespread, or
    \item prestigious, or
    \item well-documented.
\end{itemize}
This is not a finalised set of criteria, but they are under consideration.

When studying games, we aim to collect the following data (which will be feasible or infeasible to varying degrees, depending on what evidence can be found):
\begin{itemize}
    \item \textbf{Who} played a game? Was a game played throughout an entire civilisation, or primarily by the elite, commoners, adults, children, etc.?
    \item \textbf{What} was a game played with? What materials were used?
    \item \textbf{Where} was a game played (geographically)?
    \item \textbf{When} was a game played (a specific period of time)?
    \item \textbf{Why} was a game played (competitively, or to pass the time, or as a social event)?
    \item \textbf{How} was a game played (what was the rule set or were the rule sets)?
\end{itemize}
Some of the primary sources that can provide some of this data are:
\begin{itemize}
    \item \textbf{Historical texts}; in particular for rules these tend to be the most reliable sources, but they can also provide information on the other aspects.
    \item \textbf{Art}; artistic depictions of games being played can especially provide insight into the social settings in which games were played, who played them, and where they were played. Occasionally they may also provides clues for the rules.
    \item \textbf{Game boards} and other playing equipment; such objects naturally provides information about the materials that were used for game equipment. In combination with knowledge about craftsmanship, availability of materials, etc., this can provide information about aspects such as when the game was played, where it was played, by whom, etc. In rare cases, this material evidence may also contains hints about the rules (for instance if pieces are laid out in a particular configuration).
    \item \textbf{Ethnography}; this can primarily provide insight into who played (or plays) games, and in what kind of (social) context.
    \item \textbf{Archaeological context}; the context in which other sources (any of the above) are found often provides important information concerning by whom, where, when, and why games were played. For instance, a game being found in a burial site for elites may lead to different conclusions than the same game being found in a commoners' burial site.
\end{itemize}

When interpreting any of the sources described above, it is important to do so carefully and be aware of potential biases. There may be archaeological biases (e.g. some materials remain better preserved, and can therefore be more reliably discovered, than others), as well as historical biases (burial sites of elites may be better preserved than those of commoners, historical texts may include biases from their original authors, etc.). Furthermore, field work is also more common and/or thorough in certain regions than in others, and historical texts in some languages may be more easily accessible to certain researchers than others. When evidence pointing towards a certain game, playing practice, rule set, etc., is found, this also does not necessarily mean that that game, playing practice, or rule set was common. Similarly, absence of evidences does not imply absence of a game ever having existed in a certain region.




\subsection{Case Studies}

\presentation{Eddie Duggan}

In this subsection, we discuss a number of example cases that illustrate some of the uncertainties encountered when studying traditional games and how they were played. The first case is the game of \textit{Ludus Latrunculorum}. It is known that this is a board game that was played in the Roman Empire, but precise rules are not known. There is a wide variety of proposed reconstructions of its rule set \cite{Falkener1892GamesAncient, Austin1934RomanBoardGames, Murray1951, Bell1979BoardAndTableGames, Schadler1994Latrunculi, Parlett1999OxfordHistory}, inspired by fragments of evidence that have been found. In addition to the rule set(s), there has also been disagreement about where the game was played; \citeA{Falkener1892GamesAncient} argued that the game was played in ancient Egypt, but \citeA{Crist2016AncientEgyptians} identify shortcomings in this analysis and disagree.

\begin{figure}[h]
    \centering
    \begin{subfigure}{.5\textwidth}
        \centering
        \includegraphics[width=\linewidth,height=4.68cm]{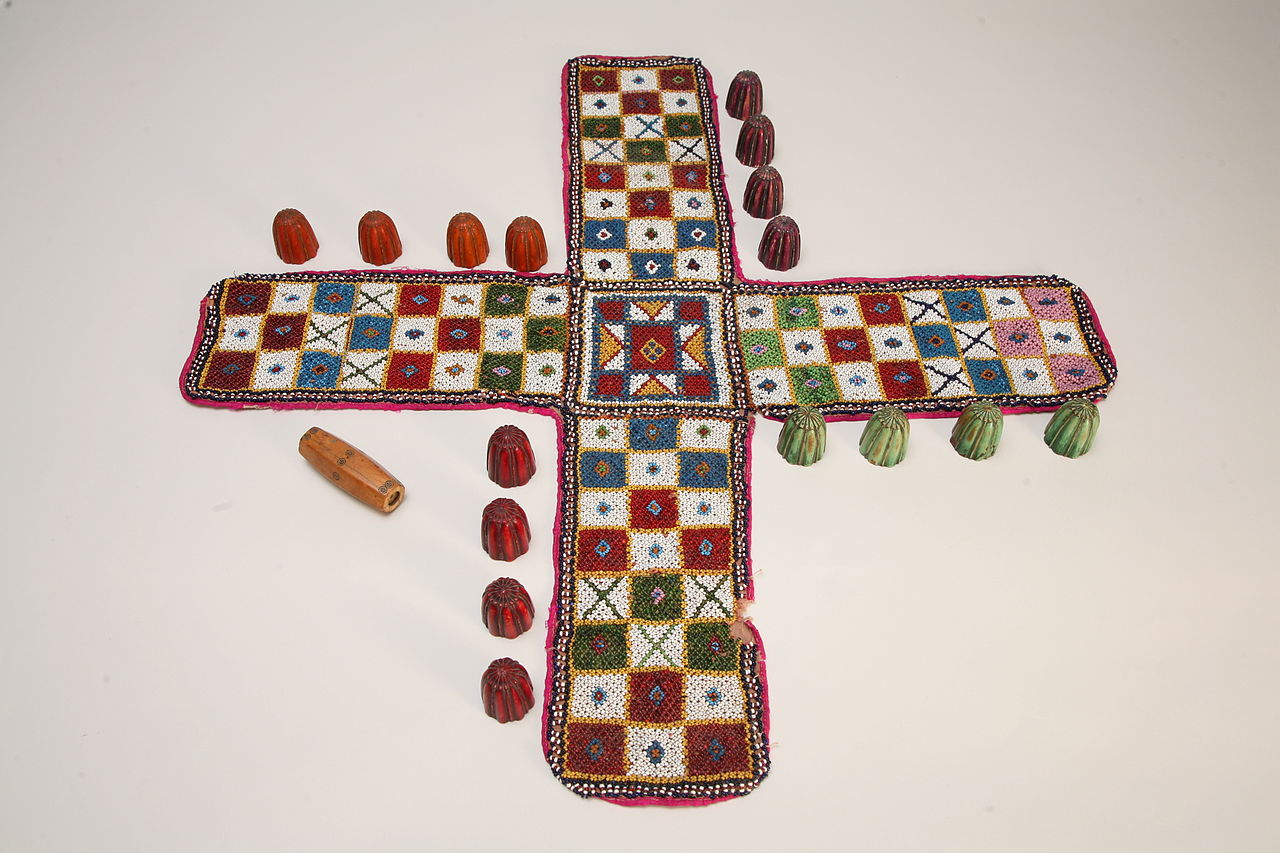}
    \end{subfigure}%
    \begin{subfigure}{.5\textwidth}
        \centering
        \includegraphics[width=\linewidth,height=4.68cm]{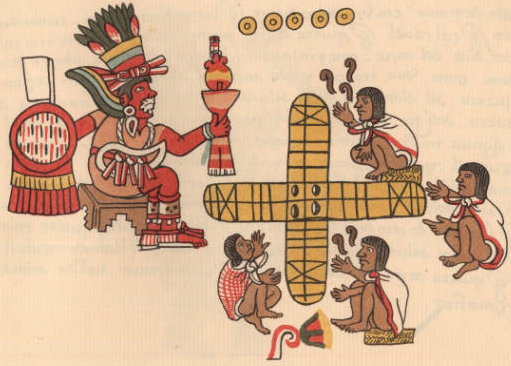}
    \end{subfigure}%
    \caption{The game of Pachisi \textbf{(left)}, and a depiction of the game of Patolli \textbf{(right)}.}
    \label{Fig:PachisiPatolli}
\end{figure}

Another interesting case is that of the Indian board game \textit{Pachisi}, and the pre-Columbian Mesoamerican board game \textit{Patolli}, which visually appear to be highly similar (see \reffigure{Fig:PachisiPatolli}). \citeA{Tylor1879Patolli} suggested this hints at pre-Columbian contact between Asia and South America, although this was later disputed by Erasmus~\cite{Erasmus1950Patolli}, who observes that the two games may have developed independently due to the principle of ``limited possibilities''. 

This is a difficult problem to assess with available techniques. 
While the games have some similarities in their rules, Caso's accounts of Patolli show that versions of the game still being played in the early 20$^{th}$ century also involved significantly  different rules~\cite{Caso1925,Caso1927}, which Thierry DePaulis cites as evidence that the games probably did not share a common ancestor. 
One of the goals of DAL is to provide empirical analyses that may shed light on such cases.

Recently, \citeA{Crist2018NearEasternGame} found evidence of the game of \textit{Hounds and Jackals} in Azerbaijan. The game is otherwise primarily known to have been played in regions such as ancient Egypt and Mesopotamia \cite{Voogt2012CulturalTransmission}. This evidence suggests that there may have been contact between these cultures (not necessarily direct contact, but possibly indirect contact through other cultures).

The rules of the \textit{Royal Game of Ur} were reconstructed from descriptions on stone tablets \cite{Finkel2007RoyalGameUr}. There is some uncertainty in regards to, for example, the tracks that players followed in this racing game. However, the set of plausible tracks can be greatly restricted by testing the playability of resulting rule sets, since some tracks would lead to unplayable games.



\subsection{Forensic Game Reconstruction} \label{SubSec:ForensicGameReconstruction}

\presentations{Fred Horn and Ulrich Sch\"adler}


Previous sections discussed how, for many games played throughout history, there are only fragments of evidence that provide some clues as to the rules, but often no complete rule sets. In this section, we provide examples of how attempts have been made at reconstructing rule sets from such evidence.

\begin{figure}[h]
    \centering
    \includegraphics[width=\textwidth]{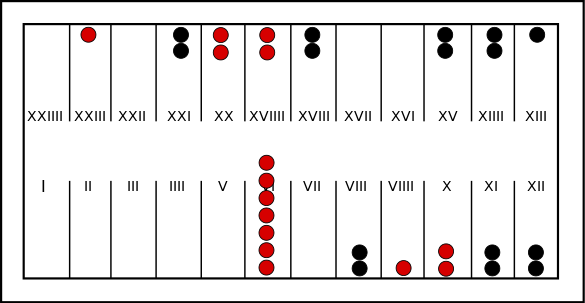}
    \caption{Game state in Emperor Zeno's game of $\uptau \acute{\upalpha} \upbeta \uplambda \upeta$ (``tabula'').}
    \label{Fig:Tabula}
\end{figure}

The primary reference to the game of $\uptau \acute{\upalpha} \upbeta \uplambda \upeta$ (``tabula'') is an epigram of Emperor Zeno, which describes a game state (depicted in \reffigure{Fig:Tabula}) in which the emperor (red in the figure) was in a strong position, but forced into a weak position due to an unlucky dice roll. The game is similar to, and likely an ancestor of, the game of Backgammon. Relevant evidence for reconstructions of the rule set that can be gathered from the epigram consists of:
\begin{itemize}
    \item Players roll three dice, which landed on a two, a five, and a six for Emperor Zeno in the game state depicted in \reffigure{Fig:Tabula}.
    \item Players are allowed to have many pieces stacked in a single space (there are seven pieces stacked in a single space in \reffigure{Fig:Tabula}).
    \item Blots (spaces occupied by only a single piece) are vulnerable to getting hit by opposing pieces (as in Backgammon).
    \item Spaces occupied by two or more pieces are not vulnerable, and there appears to be a tendency in the game to create many pairs (see \reffigure{Fig:Tabula}).
    \item The results of the three different dice must be used for separate movements (otherwise the dice roll described in the epigram would not necessarily lead to a weak position).
\end{itemize}
From this evidence, Becq de Fouqui{\`e}res created a reconstruction of the rule set in the $19^{th}$ century. 

\begin{figure}[h]
    \centering
    \includegraphics[width=\textwidth]{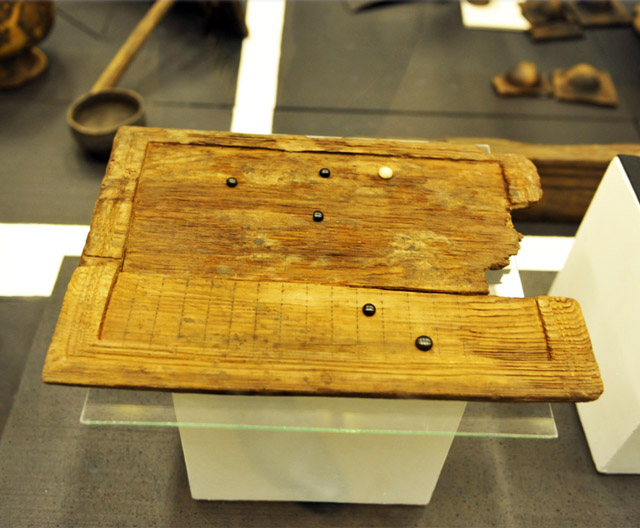}
    \caption{The Poprad game.}
    \label{Fig:Poprad}
\end{figure}

An even more complex case is that of a wooden board and Roman glass pieces found in a tomb in Poprad, Slovakia, of a Germanic chieftain who served in the Roman army \cite{Spectator2018Poprad}, as shown in Figure~\ref{Fig:Poprad}. There is some resemblance with the equipment of the game of Ludus Latrunculorum, which was played in the Roman Empire, but the board found in Poprad has significantly more squares than any other known board for that game. Only six glass pieces were found (five black and one white), which may or may not have been a complete game set. There were pieces of two different sizes, which may indicate different piece types, but may also simply be due to imperfect manufacturing. There are no other board games known to have been played during the same period of time in the same region. 

When there is very little evidence, as in the Poprad case, it should be acknowledged that the objects also may not have been used for any games at all. For example, the ``board'' that was believed to have been used to play a game referred to as ``\textit{Round Merels}'' may not have been a game at all \cite{Schadler2018RoundMerels}. Game designer and inventor Fred Horn points out a case where Thierry Depaulis sent him a picture of graffiti, and Fred Horn chose to interpret it as a board and designed rules around it. This highlights that, even when a plausible set of rules that plays well can be generated, it does not necessary imply that the material was truly used to play games at all.

It is worth clarifying here the role that Ludii might play in the task of forensic game reconstruction. 
Ludii will allow the investigator to:

\begin{enumerate}

\item model the available evidence as a partial game with known constraints, 

\item complete the rule set with ludemes deemed to be plausible based on historical and cultural context, and 

\item perform mathematical analyses of how the resulting game actually plays.

\end{enumerate} 

It is hypothesised that highly plausible rule combinations that produce highly playable games are more worthy of interest than implausible rule sets that produce poor games. 
Ludii might therefore be a useful tool for helping practitioners choose between possible interpretations of ancient evidence. 
We do not claim that it will produce a single definitive ``correct'' result for any given case, but a distribution of likely candidate solutions for the expert to assess.


\subsection{Historical Authenticity} \label{SubSec:HistoricalAuthenticity}

\presentation{Thierry Depaulis}

When attempting to reconstruct games based on partial evidence as discussed in the previous subsection, it is important to take into account the historical authenticity -- the likelihood of a combination of ludemes, or individual ludeme, occurring in a certain period of time in a certain location. Aspects that can be taken into consideration when evaluating historical authenticity include:
\begin{itemize}
    \item Other known games from the same historical context (culture, location, time, etc.). If certain properties (ludemes, or other properties emerging from combinations of ludemes) appear to be prevalent in games for which more evidence is available, it may be more likely that similar properties appear in other games for which less evidence is available.
    \item Prevalent cultures and religions within a historical context. Some properties of games tend to correlate in particular with religious aspects. For instance, stochastic elements such as dice rolls were often associated with religion or divination.
    \item Knowledge of the social settings in which games were played. For instance, sometimes games were played primarily in larger social settings, with the spectacle and betting games ``around'' a game being more important than the game itself.
    \item Knowledge of the goals with which games were played. Games were not necessarily played primarily for fun, or competition. For instance, war games may sometimes have been used to train generals, and \textit{Snakes and Ladders} was used to teach morals. Different goals for playing games may have different implications for the likelihood of certain ludemes occurring.
\end{itemize}



\subsection{Games and Mathematics} \label{SubSec:GamesAndMathematics}

\presentation{Jo\~ao Pedro Neto and Jorge Nuno Silva}

Mathematical and strategic thinking in strategy games have often been found to be closely related. Solving a mathematical problem could be viewed as a single-player game, or puzzle, in the sense that there is a fixed set of rules (e.g. axioms) that describe the legal ``moves'', and the problem is solved when a certain goal is reached using only manipulations allowed by the rules. Performance in board game competitions for children has been found to correlate with performance in math tests \cite{Ferreira2013PerspectiveOnGames}, and participating in such competitions may also subsequently improve mathematical performance in education.

Due to the similarity in abstract thinking, there may be correlations that can be found between the development of mathematical knowledge in different cultures, and the development of new types of games (or ludemes) in those same cultures. Such correlations may be discovered by annotating ludemes with underlying mathematical concepts, and comparing known data of the development of mathematics to known data of the origin of games. We note that it is likely that games and mathematical ideas -- even if they are likely to be correlated with each other in origin -- may also transfer independently of each other between cultures, which can be a complicating factor in such an analysis. An interesting observation on this topic is that the ancient Egyptians used a multiplication system reminiscent of multiplication in base $2$,\footnote{https://www.storyofmathematics.com/egyptian.html} during the same period of time in which the game of \textit{Senet} -- with binary dice -- was played.

Perhaps outside the field of \DA, a database where ludemes are annotated with underlying mathematical concepts may also be a useful tool for efforts in mathematics education, such as those described by \citeA{Ferreira2013PerspectiveOnGames}. Strategy games could be selected specifically based on the ludemes they use, in order to specifically target certain mathematical concepts to practice in education.



\section{Computational and Data-driven Techniques for \DA}

In the previous section, we illustrated some of the challenges that researchers in traditional fields such as archaeology, anthropology, history, etc. are faced with when studying how games were played throughout history, and what kind of data can be obtained and used. In this section, we describe various computational and data-driven techniques, primarily from the field of Artificial Intelligence (AI), and discuss how they may aid researchers in gaining more insight into traditional strategy games.


\subsection{GGP, GDLs and AIs}

\presentation{Abdallah Saffidine and Mark Winands}

Computer game-playing has been a topic of interest almost since the beginning of AI as a research field \cite{Shannon1950Chess}, and remained so throughout the years, with famous landmarks being the victory of \textsc{Deep Blue} \cite{Campbell2002DeepBlue} over the human world Chess champion Garry Kasparov in 1997, and the victory of \textsc{AlphaGo} \cite{SilverHuangEtAl16nature} over the 9-dan professional human Go player Lee Sedol in 2016. These landmarks are examples of research effort focussed at (approaching) perfect play. Other research goals in game-playing AI include human-level AI, human-like AI, tutoring, explainable AI players and AI for entertainment~\cite{Browne2019HumanLike}.

Programs such as \textsc{Deep Blue} and \textsc{AlphaGo} are specific to the games (Chess and Go, respectively) they have been implemented for, and cannot directly play other games. The underlying techniques and algorithms can generalise to other games \cite{Silver2018AlphaZero}, but this often requires significant engineering effort -- for instance to implement the rules of the game. To facilitate research and evaluations in more varied suites of games, Game Description Languages (GDLs) are used. These languages define formats in which game rules can be described, such that programs can be written that can run any game described in GDL. Well-known GDLs are the Stanford GDL \cite{love08} for abstract games, and the Video Game Description Language (VGDL) \cite{Schaul2013VGDL, levine_et_al:DFU:2013:4337, ebner_et_al:DFU:2013:4338} for video games. These are used in the General Game Playing (GGP) \cite{genesereth05} and General Video Game AI (GVGAI) \cite{Perez2019GVGAI} competitions, respectively.

Stanford's GDL \cite{love08} -- the current standard for GGP researchers in the types of abstract games considered in \DA\ -- is a logic-based language. An example description of the game of Tic-Tac-Toe in this language is depicted in \reffigure{Fig:GDLTicTacToe}.

\begin{figure} [h!]
\small
\fbox{%
  \parbox{.86\columnwidth}{%
  	\texttt{%
  		(role white) (role black)\\
        (init (cell 1 1 b)) (init (cell 1 2 b)) (init (cell 1 3 b))\\
        (init (cell 2 1 b)) (init (cell 2 2 b)) (init (cell 2 3 b))\\
        (init (cell 3 1 b)) (init (cell 3 2 b)) (init (cell 3 3 b))\\
        (init (control white))\\
        (<= (legal ?w (mark ?x ?y)) (true (cell ?x ?y b))\\
            \hspace*{1.5em}(true (control ?w)))\\
        (<= (legal white noop) (true (control black)))\\
        (<= (legal black noop) (true (control white)))\\
        (<= (next (cell ?m ?n x)) (does white (mark ?m ?n))\\
            \hspace*{1.5em}(true (cell ?m ?n b)))\\
        (<= (next (cell ?m ?n o)) (does black (mark ?m ?n))\\
            \hspace*{1.5em}(true (cell ?m ?n b)))\\
        (<= (next (cell ?m ?n ?w)) (true (cell ?m ?n ?w))\\
            \hspace*{1.5em}(distinct ?w b))\\
        (<= (next (cell ?m ?n b)) (does ?w (mark ?j ?k))\\
            \hspace*{1.5em}(true (cell ?m ?n b)) (or (distinct ?m ?j)\\
            \hspace*{1.5em}(distinct ?n ?k)))\\
        (<= (next (control white)) (true (control black)))\\
        (<= (next (control black)) (true (control white)))\\
        (<= (row ?m ?x) (true (cell ?m 1 ?x))\\
            \hspace*{1.5em}(true (cell ?m 2 ?x)) (true (cell ?m 3 ?x)))\\
        (<= (column ?n ?x) (true (cell 1 ?n ?x))\\
            \hspace*{1.5em}(true (cell 2 ?n ?x)) (true (cell 3 ?n ?x)))\\
        (<= (diagonal ?x) (true (cell 1 1 ?x))\\
            \hspace*{1.5em}(true (cell 2 2 ?x)) (true (cell 3 3 ?x)))\\
        (<= (diagonal ?x) (true (cell 1 3 ?x))\\
            \hspace*{1.5em}(true (cell 2 2 ?x)) (true (cell 3 1 ?x)))\\
        (<= (line ?x) (row ?m ?x))\\
        (<= (line ?x) (column ?m ?x))\\
        (<= (line ?x) (diagonal ?x))\\
        (<= open (true (cell ?m ?n b))) (<= (goal white 100) (line x))\\
        (<= (goal white 50) (not open) (not (line x)) (not (line o)))\\
        (<= (goal white 0) open (not (line x)))\\
        (<= (goal black 100) (line o))\\
        (<= (goal black 50) (not open) (not (line x)) (not (line o)))\\
        (<= (goal black 0) open (not (line o)))\\
        (<= terminal (line x))\\
        (<= terminal (line o))\\
        (<= terminal (not open))
  	}%
  }%
}
\caption{The game of Tic-Tac-Toe modelled in GDL.}
\label{Fig:GDLTicTacToe}
\end{figure}



\subsection{Ludemes and \LUDII}

\presentation{\'Eric Piette and Dennis Soemers}


The \LUDII\ program in development for the Digital Ludeme Project (DLP) uses a new Game Description Language (GDL) based on a ludeme-based approach. In contrast to the logic-based approach of Stanford's GDL, the ludeme-based approach leads to more succinct, and -- in our opinion -- more easily understandable, high-level game descriptions. For example, \reffigure{Fig:LudiiTicTacToe} depicts a description of the game of Tic-Tac-Toe in \LUDII\ (for a comparison to the same game in GDL, see \reffigure{Fig:GDLTicTacToe}). The short, relatively easily-written game descriptions in \LUDII\ make it feasible to implement the broad range of games required for \DA, which would not be feasible in a more verbose logic-based approach.

\begin{figure} [h!]
\small
\fbox{%
  \parbox{.86\columnwidth}{%
  	\texttt{%
  		(game "Tic-Tac-Toe"\\
          \hspace*{1.5em}(mode 2 (addToEmpty))\\
          \hspace*{1.5em}(equipment \{\\
            \hspace*{3em}(board (square 3) (square))\\
            \hspace*{3em}(ball P1)\\
            \hspace*{3em}(cross P2)\\
          \hspace*{1.5em}\}\\
          \hspace*{1.5em})\\
          \hspace*{1.5em}(rules\\
            \hspace*{3em}(play (to (mover) (empty)))\\
            \hspace*{3em}(end (line length:3) (result (mover) Win))\\
          \hspace*{1.5em})\\
        )
  	}%
  }%
}
\caption{The game of Tic-Tac-Toe modelled in \LUDII.}
\label{Fig:LudiiTicTacToe}
\end{figure}

Using ludemes, which often correspond to high-level game concepts, to model games is expected to be beneficial for various analyses in \DA; these are the same ludemes that may be useful to take into account in game classifications (\refsubsection{Subsec:GameClassifications}) and evaluations of historical authenticity (\refsubsection{SubSec:HistoricalAuthenticity}), to annotate with underlying mathematical concepts (\refsubsection{SubSec:GamesAndMathematics}), etc. By implementing ludemes directly in software, the system can also run games (and AIs playing them) significantly faster than the logic-based approach which requires logic-based interpreters, and AI algorithms are easier to implement without requiring highly optimised reasoners \cite{Sironi17}. Ludeme-based approaches for modelling games have also previously been shown to facilitate evolutionary generation of games \cite{browne09}. We expect this to be beneficial in the field of \DA\ for purposes such as automated reconstructions of plausible rule sets. In addition to benefits of the ludeme-based approach for applications in \DA, we expect the system to also be an interesting contribution for the game AI research community in general.

Within \LUDII, we aim to implement strong -- not necessarily superhuman, but ideally approximately human-level -- AI players for all games. This will be done using generally applicable AI techniques, such as Monte Carlo tree search (MCTS) \cite{Kocsis2006UCT, Coulom2007, Browne2012}, evolutionary methods \cite{Lucas2019EvolutionaryMethods}, and automated learning from self-play \cite{Silver2017AlphaGoZero, Anthony2017ExIt, Silver2018AlphaZero}. Those learning techniques are typically combined with compute- and data-intensive deep neural networks in attempts at reaching superhuman playing strength. We do not require superhuman playing strength, and therefore focus on linear methods which can learn significantly more quickly using similar training algorithms \cite{Soemers2019BiasingMCTS,Soemers2019LearningPolicies}. These training algorithms use simple local patterns \cite{Browne2019StrategicFeatures}, or features, as inputs. These features use the same representation across most (board) games, which allows for transferring and comparison of learned strategies across games. In the context of \DA, we aim to investigate if comparing games in terms of strategies learned within them can also help gain more insight into possible relations between games.


\subsection{Game Quality} \label{SubSec:GameQuality}

\presentations{Cameron Browne and Simon Lucas}


In this subsection, we consider the hypothesis that quantitative measures of ``game quality'' may aid the reconstruction of plausible rule sets given incomplete evidence (see \refsubsection{SubSec:ForensicGameReconstruction}). For instance, for many years the ``definitive'' rule set for Hnefatafl \cite{Murray1951} was biased to be strongly in favour of the king's side due to a translation error \cite{Ashton2010Hnefatafl}. Using modern AI techniques, such as those discussed in the previous subsection, such biases can easily be detected -- not just for a single game, but at scale, with automated tests for \textit{all} games modelled in \LUDII. 

Note that we do not claim that every rule set must result in games without clear biases in favour of some player(s) to be a plausible, historically authentical rule set; it is very well possible that games with strong biases were played. However, it can still be an indicator that a rule set may be an outlier, which could be treated as a flag that additional research -- such as double-checking the translation -- is warranted. 

Other potential indicators of game quality that may be evaluated automatically using self-play include:
\begin{itemize}
    \item Strategic depth: games that are not too easy to solve, and also not too difficult to gain some level of competence in, may be more plausible than extremely easy or extremely difficult games.
    \item Drama: games in which there is potential for a ``losing'' player to make a come-back, may have been more interesting to play and therefore have an increased likelihood of being authentic.
    \item Decisiveness: games that drag on for a long time when the outcome is clear may have been less interesting to play.
    \item Uncertainty: games in which the outcome is unclear throughout the majority of the time spent playing may have been more interesting to play.
    \item Drawishness: games that frequently end in draws may have been less interesting to play.
    \item Duration: games that take an extremely short or long time to end may be less plausible.
\end{itemize}
Again, we stress that these measures of quality are by no means requirements for games or rule sets to have been authentic, but indicators that can provide guidance for further research or point out outliers.

Throughout the majority of history, rules were passed on verbally rather than in writing. With this in mind, another aspect of ``game quality'' that may contribute to the likelihood of a rule set being authentic is the ease with which that rule set is explained and understood. Games with extremely long, detailed rule descriptions are less likely to have been played throughout history than games with rules that can easily be explained. This is not an aspect that can be evaluated automatically through self-play, but may be evaluated to some extent by looking at the ludeme-based game description.

In some cases, when there is incomplete evidence for rule sets, there may still be evidence related to some of the aspects of game quality discussed in this subsection. For instance, if it is known that a game was frequently played in large social gatherings, with a large spectacle around it, it is more likely that drama and uncertainty would be important qualities of that game.

In addition to the field of \DA, automated evaluation of some aspects of the quality of (potentially automatically generated!) games can also be a valuable contribution to game design and AI research.


\subsection{Genetics of Games} \label{SubSec:GeneticsOfGames}

\presentation{Steven Kelk}


While strategy games do not inherently have any genetic material \cite{Voogt1999DistributionMancala}, a tree of ludemes may similarly be viewed as the genotype of a game from which the way that it plays (phenotype) emerges. When games transfer to other cultures, regions, religions, parts of a population, etc., or when they are used as inspiration for the development of a new game or game variant, we often expect some of its ludemes -- though possibly not all of them -- to also transmit.

In the field of computational phylogenetics, it is assumed that populations (of organisms, or languages, or games, etc.) did not appear spontaneously, but developed through ``evolutionary'' processes which place strong constraints on the hypotheses that can plausibly explain observed similarities and differences. Computational models encoding assumptions based on those constraints can be used to construct plausible networks of ``ancestry''. In biology such networks are generally assumed to be trees, but in the case of cultural artefacts such as games we expect there to be significantly more horizontal transfer. There have been previous attempts at using phylogenetic techniques for Mancala games \cite{Eagle1999PhylogeneticMancala} and Chess-like games \cite{Kraaijeveld2001OriginChess}, but they tended to confuse the genotype and phenotype of games in these analyses \cite{Morrison2013FalseAnalogies}.

Similar to the quantitative estimates of game quality discussed in \refsubsection{SubSec:GameQuality}, we do not expect computational phylogenetics to allow for conclusions with respect to the origins of games with $100\%$ certainty. However, the resulting networks (or distributions over plausible networks) may again provide insight into plausible hypotheses to be investigated in more detail in further research. We envision outputs of computational phylogenetics to be used in combination with other analyses (such as those discussed in the previous subsection) and partial evidence from more traditional research fields (e.g. archaeology).

\subsubsection{Phylogenetic Networks}

Rather than all games coming from a single ancestor ({\it monogenesis}) it is more likely that games originated independently from multiple sources ({\it polygenesis}). This has implications for the analysis to be performed, and it is likely that we need to model a {\it phylogenetic network} of games rather than a single phylogenetic tree.  

Further, Thierry Depaulis and Alex de Voogt point out that they know of very little exchange of ideas between board games, card games and Mancala games throughout history,\footnote{Personal discussions.} suggesting that these different types of games may have developed independently of each other and that their ``family trees'' may interact only occasionally or may not overlap at all. 
However, we will include these different game types in the study, and look for interactions throughout history that may have been overlooked.


\subsection{Cultural Mapping}

\presentation{Matthew Stephenson}


One of the goals in \DA\ is to create plausible mappings of the spread of games through time, space, and cultures. Based on archaeological evidence (\refsubsection{SubSec:CorpusAndDataCollation}), such data will be available for some (variants of) games, and not available or uncertain for others. When available, such data is crucial for the evaluation of historical authenticity (\refsubsection{SubSec:HistoricalAuthenticity}) of reconstructed games (\refsubsection{SubSec:ForensicGameReconstruction}). When such historical data is not available or too uncertain, distributions over plausible explanations may be inferred from phylogenetic analysis (\refsubsection{SubSec:GeneticsOfGames}) and game reconstructions based on partial evidence and estimates of game quality (\refsubsection{SubSec:GameQuality}). Additional data will be provided by GeaCron,\footnote{\url{http://geacron.com}} which maintains a database of maps from 3000\textsc{bc} till now. This includes information on dominant cultures, religions, nations, etc. at any given geographical location at various points in time, trade routes, historical events, etc.

To facilitate the use of various data-driven techniques, a database and website are in development for the Digital Ludeme Project (DLP) which are intended to store, and provide access to, a comprehensive and reliable collection of data. To aid researchers in traditional research fields, the website will include various visualisations and search functionality for the database. All games will also be playable through \LUDII. In the future, information on these releases will be made available on \url{http://ludeme.eu/}.


\section{Conclusion} \label{Sec:Conclusion}

The aim of this inaugural international research meeting on \DA\ (DAL) was to draw together experts in its various related research fields, to promote interdisciplinary discussion in order to help define the scope and aims of this new research field, and to establish a framework for its successful continuation beyond the life of its initiating project. This gathering also doubled as the first Advisory Panel meeting of the ERC-funded Digital Ludeme Project (DLP), and was attended by almost all Advisory Panel members.

While the short format (1.5 days) did not allow much time for in-depth presentations by participants on the wide range of topics, the ensuing discussions were lively and extremely valuable in helping shape DAL, as it transitions from a set of ideas to a practical research topic, at the the first-year mark of the DLP. 
We sincerely thank all participants for their contributions to the event. 


\newpage

\begin{acknowledgements}
This research meeting was funded by the European Research Council, as part of the Digital Ludeme Project (ERC Consolidator Grant \#771292) run by Cameron Browne at Maastricht University's Department of Data Science and Knowledge Engineering. We thank Schloss Dagstuhl for facilitating the event. 
Ulrich Sch\"adler is co-investigator on the ERC-funded Locus Ludi project.\footnote{\url{https://locusludi.ch}}
\end{acknowledgements}

\begin{participants}
\participant Cameron Browne\\ Department of Data Science and Knowledge Engineering, Maastricht University
\participant Dennis J. N. J. Soemers\\ Department of Data Science and Knowledge Engineering, Maastricht University
\participant \'Eric Piette\\ Department of Data Science and Knowledge Engineering, Maastricht University
\participant Matthew Stephenson\\ Department of Data Science and Knowledge Engineering, Maastricht University
\participant Michael Conrad\\ Kunsthistorisches Institut, Universit\"at Z\"urich
\participant Walter Crist\\ Department of Anthropology, American Museum of Natural History
\participant Thierry Depaulis\\ Independent
\participant Eddie Duggan\\ Department of Science, Technology and Engineering, University of Suffolk
\participant Fred Horn\\ Independent
\participant Steven Kelk\\ Department of Data Science and Knowledge Engineering, Maastricht University
\participant Simon M. Lucas\\ School of Electronic Engineering and Computer Science, Queen Mary University of London
\participant Jo\~ao Pedro Neto\\ University of Lisbon
\participant David Parlett\\ Independent
\participant Abdallah Saffidine\\ School of Computer Science and Engineering, University of New South Wales
\participant Ulrich Sch\"adler\\ Swiss Museum of Games
\participant Jorge Nuno Silva\\ University of Lisbon
\participant Alex de Voogt\\ Economics \& Business Department, Drew University
\participant Mark H. M. Winands\\ Department of Data Science and Knowledge Engineering, Maastricht University
\end{participants}

\begin{figure*}[b]
\centering
\includegraphics[width=\textwidth]{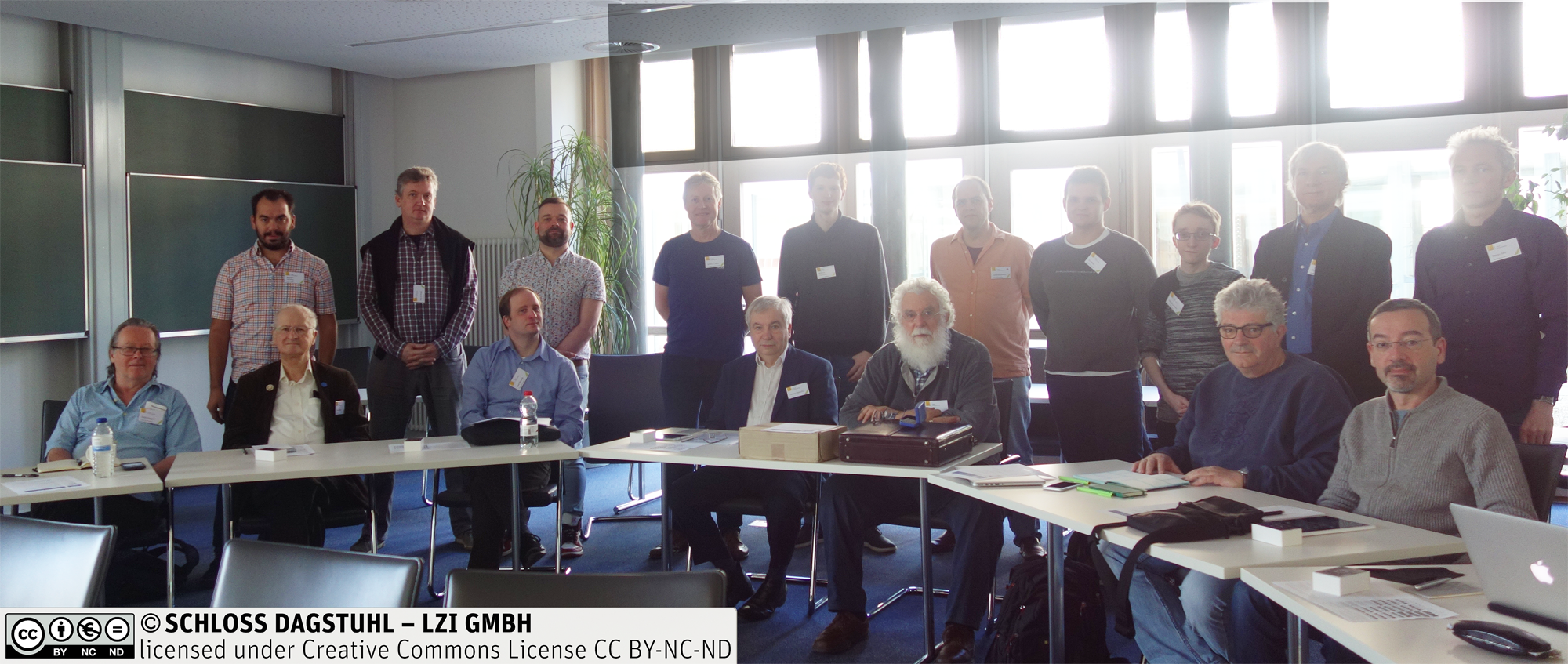}
\end{figure*}

\clearpage
\titlerunning{19153 -- \DA}
\authorrunning{Cameron Browne {\it et al.}}

\bibliography{References}

\clearpage
\titlerunning{19153 -- \DA}
\authorrunning{Cameron Browne {\it et al.}}

\appendix

\section{On the origins of the word ‘ludeme’ (French \textit{ludème})}
\textit{by Thierry Depaulis}\\
\\
It seems the word ‘ludeme’ – as \textit{ludème} (in French) – was coined in France in the early 1970s in the wave of Structuralism then intellectually dominant.
Clearly \textit{ludème} is derived from the Latin word \textit{ludus} ‘game’, with a suffix -\textit{ème} (-eme) which was much used by the structuralist school following Ferdinand de Saussure’s use of \textit{phoneme} and \textit{grapheme}. Thus we have \textit{morpheme}, then \textit{mytheme} (Lévi-Strauss), \textit{sememe}, \textit{lexeme}, and many other words ending in -\textit{eme} that describe a basic element in some sort of structure. (Most of these have been borrowed by English.)

REM. In French, words derived from Lat. \textit{ludus}, like \textit{ludique}, \textit{ludion} (‘Cartesian diver’), \textit{ludologie}, \textit{ludologue}, \textit{ludothèque} (‘toy library’), \textit{ludothécaire} (‘toy librarian’), are more common than in English.

One of the earliest occurrences of the word \textit{ludeme}, defined as a constituent of a set of rules in a game, appeared in a book about a French regional card game played in Brittany, published in 1977. The book was \textit{Anatomie d'un jeu de cartes: l'aluette ou la vache} by Alain Borvo. 
Borvo writes (p. 18): 
\begin{displayquote}
Although the goal of a general study like this one is not to provide precise rules for playing ‘Aluette’ (1), but rather to sort out the game’s principles which constitute them (what Pierre Berloquin calls ‘ludemes’)...\footnote{“Bien que le but d'une étude d'ensemble comme celle-ci ne soit pas de fournir une règle précise pour la pratique de l'aluette (1), mais plutôt de dégager des principes du jeu qui la composent (ce que Pierre Berloquin appelle les « ludèmes »)…”}
\end{displayquote}
Fn. (8) refers to sources on Aluette, among which: “Berloquin, Le livre des jeux chez Stock…”
On p. 56 Borvo applies the ‘ludemes’ theory to a detailed analysis of the rules of Aluette.

It is defined by Borvo as a ‘type rule’ (\textit{règle-type}), meaning a basic element in the ‘system’ that a game forms. The trick-taking element in a card game or the leap capture in a board game can be termed \textit{ludemes}. 

Pierre Berloquin (b. 1939) is an important French writer on games. An engineer by training he soon became interested in games and wrote some remarkable books on board and card games, with a clever classification, based on a systematic analysis of the main components.
Among his many books, there are:
\begin{itemize}
    \item \textit{Le Livre des Jeux} (1970)
    \item \textit{100 Jeux de cartes classiques} (1975)
    \item \textit{100 Jeux de table} (1976).
\end{itemize}
His Wikipédia entry is here:
\url{https://fr.wikipedia.org/wiki/Pierre_Berloquin}
(In it, Berloquin claims to have coined the word ‘ludographe’ in 1975, a word that did not been take root.)

However, the word ‘ludème’ cannot be found in all of Berloquin's classic books.
So I had a private conversation with Pierre Berloquin, whom I visited in July 2005, and who said that he probably was the man who coined the word, but could not remember when and how he had invented it. He suggested he could have used ‘ludèmes’ in his \textit{Le Livre des jeux}, but as remarked above, I could not find it.

While I reported about my conversion with Berloquin to David Parlett, David commented:
\begin{displayquote}
I am beginning to think that Borvo and Berloquin must have met each other and that Berloquin suddenly and unexpectedly coined the word ludeme in conversation and then forgot about it, whereas Borvo's brain proved fertile soil for the concept! In my \textit{Oxford Guide} I described the term ludeme as ``eminently dispensable'', but I have since come to consider this unfair, as I have subsequently found myself using it quite naturally.
\end{displayquote}

But research continued and I could read in an article by Jean-Pierre Etienvre,\footnote{Jean-Pierre Etienvre published two books about card games in the Spanish literature of the Golden Age. (They were derived from his PhD thesis.) As a linguist he studied the special lexicon of card games and was used to track the ‘earliest occurrences’ of words.} “Du jeu comme métaphore politique (à propos de quelques textes de propagande royale diffusés en Espagne à l'avènement des Bourbons)”, in \textit{Poétique}, 56 (Nov. 1983), fn. 10, p. 399: 
\begin{displayquote}
It’s Pierre Berloquin (interview in \textit{Le Monde}, 30.X.1970) who seems to be the first to speak about ludèmes. He was not followed.\footnote{“C’est Pierre Berloquin (entretien dans \textit{le Monde}, 30.X.1970) qui semble avoir parlé le premier de ludèmes. Il n’a pas fait école.”}
\end{displayquote}
Checking the archives of \textit{Le Monde} yielded this (see Appendix for the French original text):
\begin{displayquote}
CHESS, GO, WRITING\\
(...)\\
By Raphaël Sorin. Published 30 October 1970 
\end{displayquote}
(An interview with Pierre Berloquin, about his just published book \textit{Le Livre des jeux}.)
\begin{displayquote}
In your preface you raise the discussion referring to ‘ludic structure’. So your book is not a simple rulebook?\\
“I must confess that my book has also the ambition to set up the bases for a future ‘ludic world’\footnote{Here, \textit{ludique}, a feminine noun coined by Berloquin, whose meaning is unclear -- ‘a world of play’?}  and it is an essay in genetic epistemology  of games from the most simple to the most sophisticated. Sometime we will speak of ‘ludic world’ and ‘ludemes’. Until now rulebooks were empirical and essays like those of Huizingua [\textit{sic}] and Caillois ignored the actual gameplay…”
\end{displayquote}
Interestingly Berloquin gives no definition of ‘ludèmes’.

Then the word soon reappeared in a book by Pierre Guiraud, \textit{La sémiologie}, Paris, PUF, 1971 (QSJ), p. 115:
\begin{displayquote}
Semiotically, the problem of games, like that of the arts, is twofold: a morphology whose object is to reduce each game to its ``immediate constituents'' with a view to classifying them and defining their functions, that is to say, the rules of their combinations; a semantics (and a symbolism) which will have to establish the meaning and the social function of these ‘ludemes’ within a culture as well as the mythical roots which melt them and connote them.\footnote{“Sémiologiquement, le problème des jeux, comme celui des arts est double : une morphologie dont l'objet est de réduire chaque jeu à ses « constituants immédiats » en vue de les classer et de définir leurs fonctions, c'est-à-dire, les règles de leurs combinaisons ; une sémantique (et une symbolique) qui devront établir le sens et la fonction sociale de ces « ludèmes » au sein d'une culture ainsi que les racines mythiques qui les fondent et les connotent.”}
\end{displayquote}
Guiraud’s book was translated in Spanish (1972), in English (1975), and probably in other languages. Guiraud gave no reference for ‘ludemes’… He probably picked it up in Berloquin’s interview of 1970.

When, in 2011, I told Berloquin about all this he was very much surprised.
He wrote in a e-mail:
\begin{displayquote}
The reference to an interview in Le Monde in 1970 astounds me. This seems very early, but a visit to a library will give me an answer. I was close to the structuralists and I was attending seminars on structural linguistics.\footnote{“La référence à un entretien dans le Monde en 1970 me confond. Cela me semble bien tôt mais un passage dans une bibliothèque me fixera. J'étais proche des structuralistes et je suivais des séminaires de linguistique structurelle.”}
\end{displayquote}
He added that he had met Borvo one or two times. And he concluded:
\begin{displayquote}
Good gracious, the product really got unnoticed, save to Borvo,\footnote{Alain Borvo, an ethnologist and a playing-card collector, sadly, died in 2002.} if it is what he says in his Aluette.
\end{displayquote}

\section{An exerpt of \textit{Le Monde}, 30 October 1970}
(An interview with Pierre Berloquin, about his just published book \textit{Le Livre des jeux}.)

\subsection*{LES ECHECS, LE GO, L'ECRITURE}

LE\textit{ Livre des jeux} est à la fois un recueil de règles, une histoire des jeux, un discours de la méthode. L'ouvrage, aussi scientifique que les traités des théoriciens des jeux, n'est pas moins savant que les essais des sociologues du ludique. Il permet à tous de vérifier en s'amusant que la manœuvre des cartes et des pions ne ressemble à aucune autre activité humaine sinon à celles où l'on joue avec des chiffres, des bataillons et des mots. L'auteur, Pierre Berloquin, définit ci-dessous dans un entretien sa conception du jeu qui est loin pour lui d'être un loisir futile.\\
\\
Par Raphaël Sorin. Publié le 30 octobre 1970 \\
\\
• Dans votre préface, vous élevez fermement le propos en parlant de structure ludique. Votre livre n'est donc pas seulement un guide pratique ?\\
\\
``Je dois avouer que mon guide a aussi l'ambition de poser les prolégomènes à une « ludique » future et qu'il est un essai d'épistémologie génétique des jeux allant du plus simple au plus raffiné. \hl{Un jour nous parlerons de ludique et de lud{\`e}mes.} Jusqu'à présent les recueils étaient empiriques et les essais comme ceux de Huizingua [\textit{sic}] et de Caillois ignoraient la pratique même du joueur, comme s'ils s'intéressaient plus à sa subjectivité qu'à la réalité de ses efforts, tactiques et stratégiques.''\\
\\
• Les jeux sont-ils réservés à une élite cultivée seule capable de pénétrer les écrits théoriques de von Neuman et de Morgenstern ?\\
\\
``Absolument pas ! Ils sont au contraire le seul art vraiment populaire, à la portée immédiate des couches sociales les plus défavorisées. Ils n'ont pas besoin, comme les autres arts, de pénétrer d'abord dans les couches intellectuelles petites-bourgeoises. Les joueurs, sans recourir au langage ésotérique et parfois abscons de certains demi-savants, réinventent spontanément des structures intellectuelles très complexes, proches de celles des mathématiques. Caillois passe à côté de la cible quand il affirme que « c'est une des caractéristiques du jeu qu'il ne crée aucune richesse, aucune œuvre ». Chaque jeu avec son ensemble de règles est déjà une œuvre d'art, et chaque joueur est un créateur qui crée la partie qu'il est en train de jouer ; enfin il ne joue pas seulement contre son adversaire, il s'associe avec lui pour jouer contre le jeu, développer des tactiques inédites et participer ainsi à une partie invisible, plus générale et plus importante. Les joueurs luttent contre les jeux, ceux-ci meurent et sont remplacés par de nouveaux jeux au cours de ce long travail de destruction. Dans cette mesure, les jeux sont le seul exemple indiscutable de création collective.''


\end{document}